\documentclass[10pt,twocolumn,letterpaper]{article}

\usepackage[cvprfinal]{cvpr}

\usepackage{graphicx}
\usepackage{amsmath}
\usepackage{amssymb}
\usepackage{booktabs}

\usepackage[pagebackref,breaklinks,colorlinks]{hyperref}

\usepackage[capitalize]{cleveref}
\crefname{section}{Sec.}{Secs.}
\Crefname{section}{Section}{Sections}
\Crefname{table}{Table}{Tables}
\crefname{table}{Tab.}{Tabs.}

\def\cvprPaperID{931} 
\def\confName{CVPR}
\def\confYear{2023}

\usepackage{wrapfig}
\usepackage[export]{adjustbox}

\newcommand{\aaa}{{\mathbf a}}
\newcommand{\bb}{{\mathbf b}}

\newcommand{\X}{{\mathcal X}}

\newcommand{\N}{{\mathbb N}}

\newcommand{\nocontentsline}[3]{}
\newcommand{\tocless}[2]{\bgroup\let\addcontentsline=\nocontentsline#1{#2}\egroup}

\newcommand{\abs}[1]{\left\lvert#1 \right\rvert}

\newcommand{\cupdot}{\mathbin{\mathaccent\cdot\cup}}

\newcommand{\R}{\mathbb{R}}
\def\multiset#1#2{\ensuremath{\left(\kern-.3em\left(\genfrac{}{}{0pt}{}{#1}{#2}\right)\kern-.3em\right)}}

\def\be{\begin{equation}}
\def\ee{\end{equation}}
\def\beas{\begin{eqnarray*}}
\def\eeas{\end{eqnarray*}}
\def\bea{\begin{eqnarray}}
\def\eea{\end{eqnarray}}

\usepackage{mathtools}

\newcommand{\defeq}{\vcentcolon=}

\newtheorem{proofSketch}{Proof idea}[section]
\newtheorem{remark}{Remark}[section]
\newtheorem{theorem}{Theorem}[section]

\newtheorem{lemma}[theorem]{Lemma}

\newtheorem{conclusion}[theorem]{Conclusion}

\newtheorem{definition}[theorem]{Definition}

\newtheorem{proof}[theorem]{Proof}

\newtheorem{proposition}[theorem]{Proposition}

\def\rb{\right)}
\def\lb{\left(}

\def\rc{\right]}
\def\lc{\left[}

\def\rs{\right\}}
\def\ls{\left\{}

\def\sum#1#2#3{\overset{#1}{\underset{#2}{\Sigma}} #3}

\def\abs#1{\left| #1\right|}

\usepackage{mathtools} 
\usepackage{changepage} 

\usepackage{amssymb}

\newcommand{\equ}[1]{
    \begin{equation}
        #1
    \end{equation}
}

\newcommand{\eq}[1]{
    \begin{equation*}
        #1
    \end{equation}
}

\usepackage{caption}
\usepackage{subcaption}

\begin{document}


\title{Transformer Vs. MLP-Mixer: Exponential Expressive
Gap For NLP Problems}

\author{Dan Navon\\
Department of Computer Science\\
Technion, Haifa, Israel\\
{\tt\small dannav@technion.ac.il}
\and
Alex Bronstein\\
Department of Computer Science\\
Technion, Haifa, Israel\\
{\tt\small bron@cs.technion.ac.il}}

\maketitle

\begin{abstract}
    Vision-Transformers are widely used in various vision tasks.  
    Meanwhile, there is another line of works starting with the MLP-mixer trying to achieve similar performance using mlp-based architectures. Interestingly, until now those mlp-based architectures have not been adapted for NLP tasks. 
    Additionally, until now, mlp-based architectures have failed to achieve state-of-the-art performance in vision tasks. 
    In this paper, we analyze the expressive power 
    of mlp-based architectures in modeling
    dependencies between multiple different inputs simultaneously,   
    and show an exponential gap 
    between the attention and the mlp-based mechanisms.
    Our results suggest a theoretical explanation
    for the mlp inability to compete with attention-based mechanisms in NLP problems,  
    they also suggest that the performance gap in vision tasks 
    may be due to the mlp relative weakness in modeling dependencies between 
    multiple different locations,  
    and that combining smart input permutations with mlp architectures may not be enough to close the performance gap alone.
\end{abstract}

\section{Introduction}
Since ViT proposed in a seminal paper by 
Dosovitskiy et al. 
\cite{dosovitskiy2021an}  
attention-based architectures
\cite{liu2021swin, wang2022pvt, chu2021twins}
are widely used for various vision tasks,
\cite{radford2021learning, dosovitskiy2021an} 
and achieve state of the art results in 
many benchmarks including the Imagenet-1k benchmark 
\cite{ding2022davit, wortsman2022model, tu2022maxvit}.   
Bit later Radford et al. \cite{radford2021learning} 
 followed by 
 \cite{yu2022s2, lian2021mlp, guo2022hire} 
suggested that simple mlp-based models combined with 
input permutations can achieve 
similar performance for the attention-based mechanisms. 
The heart of the mlp-mixer approach is to permute the input each time before applying the mlp-layer. Their idea is that permuting the inputs would allow the mlp-based 
architecture to mix information from different tokens 
in a similar way to the attention mechanism.

It is only natural to ask whether the mlp-based 
approaches combined with some permutations  
can compete with the attention-based mechanisms also in NLP tasks.
Interestingly until now the MLP-Mixer 
have not been adapted for NLP tasks.      
Additionally, until now mlp-based approaches failed to achieve the state of the art performance on vision tasks   
but they are competitive with a small margin. 
In this paper, we seek to improve our theoretical 
understanding of the difference between the mlp 
and the attention-based architectures 
in their expressive power to model problems in 
different domains namely NLP and Vision.

We will do it by answering to some extent the following three questions 
(1) Can mlp-based models compete with the 
attention-based mechanisms also in NLP-based tasks. 
(2) Is the gap between the mlp to the attention-based mechanisms on vision tasks can be closed
or a result of a gap in the expressive power,    
and hence can not be closed without architectural 
changes. 
(3) 
What differences between NLP and Vision cause the change in the ability of MLP-models to compete with the attention-based ones between these different fields.
To answer those questions
we estimate the expressive power
of the different models   
to model connections between 
multiple variables simultaneously.    
This would allow us to compare the architecture's ability  
to compete on NLP problems   
since in NLP problems the relevant information  
does not necessarily lie in the nearest neighbors   
 and hence modeling multi-variable connections
 is necessary to get all the relevant information.
 This metric would also differentiate us 
 from the vision case where the nearest neighbors contain most of the relevant information and their small number suggests that modeling multi-variable 
connections are less important.
 
Estimating network ability to 
 model multi-variable connections   
 will require us to define some 
 metric to capture this notion, 
  and for this, we will adapt the 
  separation-rank metric 
   \cite{beylkin2002numerical, beylkin2009multivariate, cohen2017inductive}, 
   for comparing the expressive power  
   of different classes of architectures. 
   To do this we will further develop the 
   notion of the separation-rank,  
   for functions with multi-dimensional 
   range. We will define what the 
   separation-rank of a class of architectures means.
   Then we will define the notion 
   of expressive-gap between different architectural classes. 
   This expressiveness definition would capture the ability of the class architectures to model multi-variable connections  and hence to compete on NLP problems. 

  Finally, we will establish the relevant 
 bounds on the mlp and attention-based 
 architectures and will show the
 higher expressiveness of attention-based architectures relative to mlp-based architectures for NLP tasks. 
 We will also show that when fixing the parameter budget, mlp-based models have lower expressivity than transformers and that there is an exponential gap in their expressive power 
 as long as they are not able to replace 
 each multi-head-attention layer at
 the transformer with at least $1.58$ mlp-layers. 
 This means that mlp-based models 
 should be significantly deeper, to 
 achieve the same level of expressiveness.
 
Using our theory we will suggest  
 theoretical answers to the above three 
 questions. 
 (1) Since mlp-based models 
 are significantly less expressive
 in their ability to model multi-variable connections,  
  we will suggest that they are not fitted for 
  NLP problems,  
  including the mlp-mixer-based architectures.
  (2) Since in vision also it is reasonable that there is some importance in modeling multi-variable connections, we suggest it as a possible reason for the existing gap between the attention-based and the mlp-based architectures in vision tasks.
  (3) As for the difference between the NLP 
  and the vision tasks, our results suggest 
  that the mlp-based architectures,  
  may be competitive for vision tasks   
  due to the lower importance of modeling 
  the multi-variable connections there  
  and the higher importance of the nearest neighbors  
  and their limited number. 
  In NLP however, this is no longer true    
  and mlp-based models would no longer be 
  expressive enough to obtain competitive results.
  
 Using our theory we predict 
  bounds on the optimal depth-to-width ratio 
  for mlp-mixer models. These bounds are different from the bounds for transformer architectures. 
  We will test our predictions   
  by comparing the accuracy of mlp-mixer models 
  with a varied depth-to-width ratio 
  on a variety of vision and NLP datasets.
  We further predict that mlp-mixer, due to its 
  weaker expressive power, 
  would require longer training, 
  and larger data size to decrease the gap, as seen in many cases when training
  models with the same architecture
  but different budgets 
  that larger models tend to converge faster.
  And assess these predictions by the experiments 
  reported by Tolstikhin et al. \cite{tolstikhin2021mlp}.
 
 To sum up, our contributions using an exact mathematical analysis
 we show an exponential gap in expressive power   
 between mlp-mixer and attention based-architectures.   
 Our results show the expressive weakness of mlp and mlp-mixer 
 architectures, for NLP problems, and suggest that also for 
 vision problems mlp-based architectures, including mlp-mixer, 
 are weaker in modeling complicated connections between 
 multiple variables simultaneously. 
 We extended the separation-rank
 definition further into the multi-dimensional and the class 
 of architectures cases, and define formally  
 how to compare the expressive power between different 
 architectural classes in terms of the separation-rank metric. 
 Finally, we establish a few basic lemmas about the separation 
 rank properties and came up with a new way to bound 
 the separation rank of complicated deep learning architectures in a recursive way.

\section{Related works}
Modeling in computer vision has long been dominated
by convolutional neural networks (CNNs). Beginning with
AlexNet \cite{NIPS2012_c399862d} and its revolutionary performance on the ImageNet image classification
challenge. CNN architectures have evolved to become increasingly powerful through
greater scale \cite{7780459, zerhouni2017wide},
more extensive connections 
\cite{huang2017densely}, and more sophisticated forms of convolution 
\cite{8237351, xie2017aggregated, zhu2019deformable},  
with CNN serving as the backbone networks 
for a variety of vision tasks. 
These architectural advances have led to performance
improvements that have broadly lifted the entire field. 
On the other hand, the evolution of network architectures
in natural language processing (NLP) has taken a different
path, where the prevalent architecture today instead is the
transformer \cite{NIPS2017_3f5ee243} designed for sequence modeling and
transduction tasks. The transformer is notable for its use
of attention to model long-range dependencies in the data.  
Its tremendous success in the language domain has led researchers to investigate 
its adaptation to computer vision, 
where it has recently demonstrated promising results on certain tasks, specifically image classification \cite{dosovitskiy2021an},
and joint vision-language modeling \cite{radford2021learning}.

There is another line of works, started by \cite{radford2021learning}, 
trying to improve the mlp-based architecture for vision 
purposes. Existing MLP-like models share a similar macro framework, but have different block designs, MLP-like models usually divide one input image into patches
like in vision transformers, and then perform two  main steps, 
especially token-mixing steps are different from the existing methods. 
ViP \cite{hou2022vision} mixes information along the height and width dimensions,   
by summing permutations on those dimensions 
before applying the mlp layer, 
S2-MLP \cite{yu2022s2} uses another spatial shift permutation step to enable information interaction among tokens, Hire-MLP permutes tokens within a local region and cross local regions, and in common all of these MLP-like methods rely on permutation matrices followed by the linear operator.
 
The current state of the art, however,
is achieved by attention-based models, 
and although when training on large-scale data-sets, such as JFT-300M \cite{sun2017revisiting},
MLP-mixer attains similar accuracy when moving into 
medium-scale data-sets such as ImageNet-1k there is a
clear performance gap.
Specifically, Mixer-Base-16
\cite{heo2021rethinking}
achieves only a $76.44$, whereas ViT-Base-16 \cite{dosovitskiy2021an} achieves a $79.67$. 

The research on the expressive power of NN has a long history, 
in 2016 Cohen and Shashua \cite{cohen2016inductive} introduced 
the separation-rank metric, to quantify the expressive power of CNNs  
and to mathematically quantify the difference between vision and NLP 
 that creates the relative success of CNNs in vision vs. NLP. 
This work has started a line of works 
Cohen et al. \cite{cohen2017analysis}, 
Levine et al. \cite{levine2020limits}, 
Weiss et al. \cite{wies2021transformer}
that use and develop those tools to mathematically quantify the effectiveness of different NN architectures and training regimes \cite{levine2021inductive}. 
In this work, we continue this line of work further  by comparing the expressive power of transformer and mlp-like architectures in modeling multi-variable dependencies.
Our results show the superiority of attention-based architectures in modeling such dependencies.

\section{Problem formulation}\label{section4}
In this section, we will present a formal definition
for the MLP-mixer architecture   
followed by some relaxations on the analyzed 
models, during our analysis we will use the 
$\sigma_2\lb x\rb = \lb ABS\lb x\rb\rb^2\,$ 
activation as a relaxing assumption, 
we will justify this assumption later in 
this section (\ref{mlp_relax}).

\subsection{MLP-mixer formulation}
\begin{definition}
    Let $\,y_{p}^2\,$ be a fully connected network  
    with residual connections, depth $p$ and $\sigma_2$ activation. 
    Then it can be written as
    $y_p^2
    = L_p^2 \circ ... \circ L_1^2\lb X\rb$   
    where $L_i^2$ denotes the $i$ layer  
    and can be written as 
    \equ{
        L_i^2\lb X\rb = \sigma_2\lb W_i X\rb \,+\, \mathbf{1}_R\lc i\rc \, X
    }
    where $R \subset \lc m\rc$ 
    is the set of the indices of all the layers with residual 
    connections.  
\end{definition}

The MLP-mixer is defined by applying a linear layer 
on the rows and the columns iteratively. This can be 
formulated as transposing the input before each even layer, 
as done in the following definition 

\begin{definition}\label{def16_2}
    Let 
    $y_{p, m, n}^{MM}: \mathbb{R}^{n \times m}
    \to \mathbb{R}^{n \times m}$ 
    be an MLP-mixer architecture 
    with residual connections no normalization layers and with $\sigma_2\lb x\rb = x^2$ activations.
    Then, it can be written in the form 
    $\;y_{p, m, n}^{MM}\lb X\rb = 
    L_p^{2, MM} \circ \, .\,.\,.\, \circ L_1^{2, MM}\lb X\rb\;$, 
    where
    $\;L_i^{2, \,MM}:\, \mathbb{R}^{n \times m} \to  \mathbb{R}^{n \times m}\;$ denotes the $i$ layer and 
    is defined by
    \begin{align}
        &
        L_i^{2, \,MM}\lb X\rb 
         = 
        \mathbf{1}_O\lc k\rc \cdot
        \sigma_2\left(W_k^o X\right)
        \quad\quad\quad\quad\quad\;\;\;
        \label{eq162}\\&\quad\quad\quad\quad\quad\;
        + \mathbf{1}_E\lc k\rc 
        \cdot \sigma_2\lb X W_k^e  \rb 
        + \mathbf{1}_R\lc k\rc X
        \notag            
    \end{align}
    
    where $X$ is the input, and $W_k^o$ is the weights matrix 
    when $k$ is odd,
    while $W_k^e$ is the weights matrix where $k$ is even. 
    More formally,  
    $\;X \,\in\, \mathbb{R}^{n \times m}\;$
    while 
    $\;W_k^o,\, W_k^e 
    \,\in\, \mathbb{R}^{m \times n}\;$,  
    where $E$ and $O$ are the sets of even and odd indices correspondingly 
    i.e $\;E \defeq 2\mathbb{N} \cap \lc p\rc\;$
    while 
    $\;O \defeq \lb 2 \mathbb{N} +1\rb\cap \lc p\rc\;$. 

    More generally, if more general permutations are combined, 
    which are not necessarily transposes, then a more general formulation would be
    \begin{align}
        &
        L_i^{2, \,MM}\lb X\rb 
         = 
        \mathbf{1}_O\lc k\rc \cdot
        \sigma_2\lb W_k^o \pi_o\lb X\rb\rb
        \quad\quad\quad\quad\;\;
        \label{eq7_2}\\&\quad\quad\quad\quad\quad\,
        + \mathbf{1}_E\lc k\rc 
        \cdot \sigma_2\lb \pi_e\lb X\rb W_k^e  \rb 
        + \mathbf{1}_R\lc k\rc \pi_r\lb X\rb
        \notag            
    \end{align}
    
    Where $\pi_o,\, \pi_e, \, \pi_r \in S_{n \cdot m}$ are permutations over the input matrix elements,   
    and $R \subseteq \lc p\rc$ is the subset containing the indices of all the layers with residual connections.
    \end{definition}

    \begin{remark}
        In the last definition (\ref{def16_2})  
        the first equation captures only the 
        MLP-mixer properties (\ref{eq162}),  
        while the second equation (\ref{eq7_2}) intended to capture the properties of some of the variants like the model described in \cite{yu2022s2}. 
        It of course captures also the original MLP-mixer properties, 
        since it can be that $\pi_e = \pi_o = \pi_R = e$, 
        where $e$ is the identity element of $S_{m \cdot n}$. 
        Hence we would refer to equation (\ref{eq7_2}) when talking about MLP-mixer from here on since it is more general.
    \end{remark}
    
    \begin{remark}
        Although equation (\ref{eq7_2})  
        is intended to capture some more variants, 
        it still does not captures all of them,  
        like the variant introduced in \cite{hou2022vision}  
        which sums up a few different permutations each time before applying the mlp. However, it did capture the essence,    
        and the proof can be extended also 
        for those more sophisticated variants.
    \end{remark}

\subsection{Relaxing assumptions}
In this subsection, we will state some relaxations 
on the analyzed models  
that would make our analysis simpler,  
while preserving the validity 
of our findings at the same time.

\textbf{Transformer relaxations.}
Following 
\cite{levine2021inductive, levine2020depth, wies2021transformer}
we will assume that all the mlp layers are at the end,
will remove all the normalization layers, 
and omit the ReLU and softmax non-linearities.
 We refer the reader to 
  Levine et al. \cite{levine2020depth}, 
  Wies et al. \cite{wies2021transformer}
for a discussion on the impact of these relaxations. 
Essentially, they are shown to weaken the overall
network power but still allow a meaningful comparison of the self-attention integration abilities.

However, in this work  
our main goal is to lower bound 
the transformer expressivity,
and show that this lower bound is still higher than the appropriate upper bound we establish for the mlp-based architectures. 
Hence analyzing a weaker version of the transformer, 
 and showing that even this weaker version 
 is stronger than the mlp-based architectures,  
 doesn't weaken our results.


\textbf{Mixer relaxations.}\label{mlp_relax}
For easiness of our analysis  
we will assume the  
$\sigma_2\lb x\rb=\lb ABS\lc x\rc\rb^2$
 activation. 
 Notice that 
 the $\lb ABS\lc x\rc\rb^2$ activation is 
 universal from the universality of $ABS\lc x\rc$  
 \cite{chatziafratis2019depth} 
 and that assuming positivity does not
 affect the network information mixing 
 abilities as measured by the separation-rank metric \cite[p.~14]{levine2021inductive}. 
Further justification for this relaxation is provided by the first experiment (\ref{exp_1}).



\section{Separation-rank}
\subsection{Introducing the separation rank}\label{sec:2:2:1}
	The separation rank, introduced in \cite{beylkin2002numerical} for high-dimensional numerical analysis, 
	was employed for various applications, \eg,~chemistry~\cite{harrison2003multiresolution}, particle engineering~\cite{hackbusch2006efficient}, and machine learning~\cite{beylkin2009multivariate}. 
	More recently,
	the separation rank has been established as a measure of dependencies modeled by deep convolutional and recurrent networks \wrt~their inputs~\cite{cohen2017inductive,cohen2017analysis,levine2018benefits}. 
	more recently, \cite{levine2020limits,wies2021vocabulary} employed this measure for studying the expressivity of a self-attention architecture with respect to its input. 
	
	For a function $y(A,B)$ over variables $A=\{\aaa^j\in\X\}_{j=1}^M$ and $B=\{\bb^j\in\X\}_{j=1}^M$, the separation rank \wrt~$(A,B)$ is the minimal number of summands that together sum up to equal $y(A,B)$, where each summand is \emph{multiplicatively separable \wrt~$(A,B)$}, \ie,~is equal to a product of two functions~--~one that intakes only $A$ variables and another that intakes only $B$ variables. 
	Formally the \emph{separation rank} of 
	$y:\X^{2M}\to\R$ \wrt~$(A,B)$ is defined as:
	\begin{equation*}
    sep_{\lb A,B\rb}\lb y\rb 
    := 
    \underset{R\in\N}{\min}\ls
        R \;\Big{|}\; 
        \begin{split}
            &
            \exists\lb g_i\rb_{i=1}^R,\, 
            \lb g'_i\rb_{i=1}^R: 
            \mathcal{X}^M \to \mathbb{R} 
            \\ & 
            y\lb A,B\rb
            = 
            \overset{R}{
            \underset{r=1}{\Sigma}}\; 
            g_{r}\lb A\rb\cdot 
            g'_r\lb B\rb  
        \end{split}
        \rs
    \end{equation*}

	If the separation rank of a function \wrt~$(A, B)$ is~$1$, the function is multiplicatively separable \wrt~$(A, B)$, meaning it cannot take into account consistency between $A$ and $B$.
	In a statistical setting, if~$y$ is a probability density function, this would mean that $A$ and $B$ are statistically independent.
	The higher $sep_{\lb A,B\rb}\lb y\rb$ is the farther is~$y$ from this situation, \ie,~ the more it models dependency between $A$ and $B$.
	
	\subsection{Extending the separation rank}
	In our case, 
	we have the architecture 
	$y_{\Theta}:\mathbb{R}^{k \times l} \to \mathbb{R}^{m \times n}$ with $\Theta$ as the parameters. 
	We will denote by $y_\Theta$ a transformer architecture  
	and by $y_\Theta^{MM}$ an mlp-based architecture.
	The architecture output is given in matrix form, and we are interested in measuring the ability of our network to model dependencies between different locations of the input. 
	As we move further into NLP tasks, the connections we will be interested in modeling will be connections between multiple different and not necessarily close positions. Hence we will adopt a balanced partition of the inputs, i.e we would take $\abs{A}=\abs{B}$, and then $sep_{\lb A, B\rb}\lb y\rb$ 
	will just measure the ability to model connections between 
	different places at the input that are not necessarily close to each other.
	
	Finally, it is shown at \cite{levine2022tensors} that for transformer architecture
	$sep_{\lb A, B\rb}\lb y\rb$ is invariant under the different balanced partitions. However, this property may not be true when handling mlp-based architectures. Hence we will define the supermom-separation-rank to be the maximal separation rank an architecture can achieve 
	relative to some balanced partition, i.e
	$sup-sep\lb y\rb \,=\, 
	\underset{ A\cupdot B = \lc 2m\rc}{\sup}\, sep_{\lb A, B\rb}\lb y\rb$. 
	Similarly, we will define the infimum-separation-rank  
	to be the infimum separation rank the architecture can get relative 
	to some balanced partition, i.e 
	$\;inf-sep\lb y\rb \,=\, \underset{ A\cupdot B = \lc 2m\rc}{\inf}\, 
	sep_{\lb A, B\rb}\lb y\rb$.
	
	However, since we are dealing with multidimensional architectures  
	we will expand our definition 
	further into the multidimensional case. 
	Denote by  
	$y:\mathcal{X}^{2m} \to \mathbb{R}^{n \times m}$ 
	a multi-dimensional architecture,  
	we will define the supremum-separation-rank as  
	$sup-sep\lb y\rb =
	\underset{i,j\in \lc n\rc \times \lc m\rc}{\sup} sup-sep\lb y_{i, j}\rb$. 
	Similarly, the infimum-separation-rank would be extended for the multi-dimensional 
    case to be the minimal inf-sep-rank achieved by some of the components,  and more formally
    $inf-sep\lb y\rb =
	\underset{i,j\in \lc n\rc \times \lc m\rc}{\inf} sup-inf\lb y_{i, j}\rb$ . 
    
    \subsection{Expressive gap definition}
    In this subsection, we will define 
    how to compare the  
    expressive power of different architectures  
    using the $inf-sep\,$ and $sup-sep\,$ defined thus far. 
    So denote by 
    $y_{1, \Theta},\, y_{2, \Theta}: \mathcal{X}^{2 m} \to \mathbb{R}^{n \times m}\,$ two  
    architectures with $\Theta\,$ as learned parameters,  
    we will say that $y_{2,\, \Theta}\,$ 
    is more expressive than $y_{1,\, \Theta}\,$, 
    if 
    $\,sup-sep\lb y_{1,\, \Theta}\rb
    \,<\, 
    inf-sep\lb y_{2,\, \Theta}\rb
    \,$. 
    Similarly,  
    we will say that $y_{2,\, \Theta}\,$ 
    is asymptotically more expressive than 
    $y_{1,\, \Theta}\,$ 
    and will denote it by 
    $y_{1,\, \Theta} \,\prec\, y_{2,\, \Theta}\,$
    if  
    $\underset{\abs{\Theta}\to \infty}{\lim}\;
    \frac{inf-sep\lb y_{2,\, \Theta}\rb}
    {sup-sep\lb y_{1,\, \Theta}\rb} = \infty\,$ holds, 
     when $\abs{\Theta}$ denotes the number of parameters. 
    Assuming further that the depth is varied we 
    will compare the expressiveness as follows
    \begin{definition}
        Let 
        $y_{1,\,\Theta}^p,\; y_{2,\,\Theta}^p:\mathcal{X}^{2m}\to \mathbb{R}^{n\times m}$ 
        be two architectures  
        with parameters $\Theta\,$  
        and budget dependent 
        architectural parameter $p\,$, 
        let's say the depth of the network.  
        Assume further that there is some 
        monotone increasing function 
        $f: \mathbb{N} \to \mathbb{R}\,$ 
        with $\,\underset{p \to \infty}{\lim} f\lb p\rb = \infty\,$
        s.t. 
        $\underset{\abs{\Theta}\to \infty}{\lim}\;
        \frac{\log\, inf-sep\lb y_{2,\,\Theta}\rb}
        {\log\, sup-sep\lb y_{1,\,\Theta}\rb}$ 
        is going to $\infty\,$ faster than $f\lb p\rb$, 
        and more formally 
        $\underset{p \to \infty}{\lim} f\lb p\rb = \infty$ and
        $\frac{\log\, inf-sep\lb y_{2,\,\Theta}^p\rb}
        {\log\, sup-sep\lb y_{1,\,\Theta}^p\rb}
         = 
        \Omega\lb f\lb p\rb\rb\,$.  
        Then we would say that $y_{2,\,\Theta}^p\,$ is f-asymptotically more expressive
        than $y_{1,\,\Theta}^p\,$, 
        and will denote it by 
        $y_{1,\,\Theta}^p \prec_f y_{2,\,\Theta}^p$. 
    \end{definition}

    Finally denoting by 
    $\mathcal{F}_B =
    \ls y_{\Theta}^p 
    \,|\; p\in P \;\;\wedge\;\; \abs{\Theta} \leq B\rs$ 
    a class of architectures with budget $B$ and 
    architectural parameters $p$,
    where $p$ is the parameters of the architecture shape,  
    like the depth-to-width ratio, the embedding dimension, and the number of heads. We will define the separation rank of the class $\mathcal{F}_B$ 
    as the separation rank the wisest architectural parameters choice can give to us within the class, i.e 
    $sep\lb \mathcal{F}_B\rb 
    = \underset{p\in P}{\sup}\; sep\lb y_{\Theta}^p\rb$. 
    Similarly, for the supremum and the infimum separation ranks,  
    we would have 
    $sup-sep\lb \mathcal{F}_B\rb 
    = \underset{p\in P}{\sup}\; sup-sep\lb y_{\Theta}^p\rb$ 
    and 
    $inf-sep\lb \mathcal{F}_B\rb 
    = \underset{p\in P}{\sup}\; inf-sep\lb y_{\Theta}^p\rb$.
    And exactly like in the case of architecture, we will 
    define the dominance between classes of architectures as follows: 
    \begin{definition}
        Let 
        $\mathcal{F}_{B,\, P},\; \mathcal{G}_{B,\, P}$ be two different classes of architectures   
        we say that $\mathcal{F}_{B,\, P}$ is asymptotically more expressive then 
        $\mathcal{G}_{B,\, P}$, and will denote it by 
        $\mathcal{G}_{B,\, P} \prec \mathcal{F}_{B,\, P}$,
        if  
        $\underset{B \to \infty}{\lim}\,
         \frac{\log\, inf-sep\lb \mathcal{F}_{B,\, P}\rb}{\log\, sup-sep\lb \mathcal{G}_{B,\, P}\rb} = \infty
        $, where $B$ denotes the number of parameters. 
        If furthermore, there exist some budget dependent  
        architectural parameter $p$,
        let say the depth of the network as a function of the parameters budget, s.t 
        $\underset{B\to \infty}{\lim}\;
        \frac{\log\, inf-sep\lb \mathcal{F}_{B,\, P}\rb}{\log\, sup-sep\lb \mathcal{G}_{B,\, P}\rb}\,$ 
        is going to $\infty$ faster then $f\lb p\rb$,   
        and more formally 
        $\underset{p \to \infty}{\lim} f\lb p\rb = \infty$ and
        $\frac{\log\, inf-sep\lb \mathcal{F}_{B,\, P}\rb}{\log\, sup-sep\lb \mathcal{G}_{B,\, P}\rb}
         = \Omega\lb f\lb p\rb\rb$.  
        Then we would say that the class $\mathcal{F}_{B,\, P}\,$ is 
        $f$-asymptotically 
        more expressive
        than the class $\mathcal{G}_{B,\, P}\,$, and will denote it by 
        $\mathcal{G}_{B,\, P} \prec_f \mathcal{F}_{B,\, P}\,$. 
    \end{definition}

\section{Separation-rank upper-bounds}

In the following subsections, we will develop tools for proving the following theorem which is also the main result of this paper

\begin{theorem}
    Let $\mathcal{F}_{B, p}^{T}\,$ be the class 
    of all the transformers architectures    
    with up to $B\,$ parameters and depth $p\,$,   
    and let $\mathcal{F}_{B, p}^{MM}\,$ be the class of all the mlp-architectures, 
    possibly with permutations of the input before each mlp-layer,  
    and with up to $B\,$ parameters and depth $p\,$.
    Then we have the following asymptotic relation 
    $\frac{\log\, inf-sep\lb \mathcal{F}_{B}^{T}\rb}{\log\, sup-sep\lb \mathcal{F}_{B}^{MM}\rb} = \Omega\lb \lb \frac{3}{2}\rb^{p}\rb\,$, 
    and more formally we have 
    $\mathcal{F}_{B, p}^{MM} 
    \prec_{\lb \frac{3}{2}\rb^p} 
    \mathcal{F}_{B, p}^{T}\,$.
\end{theorem}

\begin{proofSketch}
    In the proof, we upper bound 
    $sup-sep\lb \mathcal{F}_{B, p}^{MM}\rb\,$
    while lower bounding 
    $\,inf-sep\lb \mathcal{F}_{B, p}^T\rb\,$.  
    Then, we compare these two bounds asymptotically to get the desired asymptotic relation.  
    The lower bound is obtained mainly, 
    by relying on a similar lower bound taken from theorem 7.1 at \cite{levine2022tensors}.
    While upper bounding $sup-sep\lb \mathcal{F}_{B, p}^{MM}\rb$, is obtained by using a recursive argument of bounding the $sep-rank$ of all of the small components of the network first. 
    Then recursively bound the $sep-rank$ of larger and larger components  
    until we reach a bound for all of the architecture. 
    $\hfill\square$
\end{proofSketch}

\textbf{Elementary operations bound.}
We will start with some simple lemmas  
about the behavior of the $sep-rank$  
under the basic operations involved in each layer. 

Being more formal, let 
$f, g: \mathbb{R}^{k \times l} \to \mathbb{R}^{n \times m}$ and 
$h: \mathbb{R}^{n \times m} \to \mathbb{R}^{r \times s}$ be 
matrix functions, where $h$ is some function of $f,\, g$,
 and we want to bound the separation rank of $h$, 
 i.e $sep-rank\lb h\rb$ in terms of $sep-rank\lb f\rb$ 
 and $sep-rank\lb g\rb$. 
 Specifically, the $h$ of interest for us are the basic operations involved in the network definition, or just
\begin{align*}
    & 
        \sigma_2\lb X\rb,\;
        f\lb X\rb \odot g\lb X\rb, \;
        f\lb X\rb \,\cdot\, g\lb X\rb,\;
    \\ & 
        f\lb X\rb + g\lb X\rb, \;
        W f\lb X\rb,\;
        f \circ g\lb X\rb     
\end{align*}
And for each such form of $h$-function  
we will establish a bound on $sep-rank\lb h\rb$ of the form
\begin{align*}
    sep-rank\lb h\rb 
    \leq 
    \phi
    \lb sep-rank\lb f\rb,\, 
    sep-rank\lb g\rb\rb
\end{align*}
when $\phi:\mathbb{N}^2 \to \mathbb{N}$ is  scalar function. 
All these bounds are proved in the appendices, 
and result in the following sequence of upper bounds: 

\begin{lemma}\label{lem_last_1}
    Let $\;f,\; g: \mathbb{R}^{k \times l} \,\to\, \mathbb{R}^{n \times m}\;$ 
    be a matrix function, and let $\;k_f \defeq sep-rank\lb f\lb X\rb\rb\;$
    be the separation-rank of $f$. Then we have the following properties
    \begin{adjustwidth}{.3cm}{}
        \begin{description}
           \item[(i)] Separation rank is a sub-additive operator
           \begin{align}
            &
                sep-rank\lb f\lb X\rb + g\lb X\rb \rb 
                \label{lem24}\\ & \quad\;\;\;
                \leq 
                sep-rank\lb f\lb X\rb\rb  
                + sep-rank\lb g\lb X\rb \rb   
                \nonumber
            \end{align}
            \item[(ii)] Separation-rank is invariant under permutations.
            More formally,   
            let $\pi \in S_{n \cdot m}$ be a permutation over the entries of 
            $\;n \times m\;$ matrices, and let $f: \mathbb{R}^{k \times l} \to \mathbb{R}^{n \times m}$ be a matrix function. 
            Then the following equality holds  
            \equ{
                sep-rank\lb \pi \circ f\lb X\rb\rb = sep-rank\lb f\lb X\rb\rb 
            }
            \item[(iii)] \label{lem218}
            For 
            $Id: M_{n \times m}\lb \mathbb{R}\rb 
            \to M_{n \times m}\lb \mathbb{R}\rb$ 
            we have
            \equ{
                sep-rank\lc Id\lb X\rb\rc \leq 2 
            }   
            \item[(iv)] \label{lem22} 
            The following inequality holds
            \equ{
                sep-rank\lb f\lb X\rb^{\odot 2}\rb
                \,\leq\,  
                \binom{k_f+1}{2}
            }
            where $\odot$ is the Hadmard product and is defined by  
            $\,\lb A^{\odot k}\rb_{i j} = \lb A_{i j}\rb^k$.
        \end{description}
    \end{adjustwidth}
\end{lemma}

\begin{proof}\label{lem_last_1}
    We will bring the proof of clauses (i), (ii), (iii) here, 
    the proof of clause (iv) is left for the appendices
    \begin{adjustwidth}{.3cm}{}
        \begin{description}
        \item[(i)] \label{proof_216}
        Denoting 
        $k_f\defeq sep-rank\lb f\rb$
        and using the $sep-rank$ definition there exists 
        $\lb f'_1 \rb_{i=1}^{k_f}, \lb f''_1 \rb_{i=1}^{k_f}$ 
        s.t. 
        \begin{align*}
             f\lb X\rb
              = 
             \overset{k_f}{\underset{i=1}{\Sigma}}\;  
             f'_i\lb X_A\rb \odot f''_i\lb X_B\rb   
        \end{align*}
        similarly denoting $k_g\defeq sep-rank\lb g\rb$
         there exist 
        $\lb g'_1 \rb_{i=1}^{k_f}, \lb g''_1 \rb_{i=1}^{k_f}$
        s.t. 
        \begin{align*}
            g\lb X\rb = \overset{k_g}{\underset{i=1}{\Sigma}}\;  g'_i\lb X_A\rb \odot g''_i\lb X_B\rb
        \end{align*}
        In particular
        \begin{align*}
            & 
            f\lb X\rb + g\lb X\rb
             = 
            \overset{k_f}{\underset{i=1}{\Sigma}}\;  
            f'_i\lb X_A\rb \odot f''_i\lb X_B\rb
            \\ & 
            \quad\quad\quad\,
            \quad\quad\quad\;
             + 
            \overset{k_g}{\underset{i=1}{\Sigma}}\; 
            g'_i\lb X_A\rb \odot g''_i\lb X_B\rb
        \end{align*}
        Is valid decomposition for $f+g$ with $k_f + k_g$ elements, and hence we have 
        \begin{align*}
            & 
            sep-rank\lb f\lb X\rb + g\lb X\rb \rb 
            \leq k_f + k_g 
            \quad\quad\quad\quad
            \\ & \quad\quad\;
            = sep-rank\lb f\lb X\rb\rb  + sep-rank\lb g\lb X\rb \rb
        \end{align*}
        As needed  $\hfill\square$
        \item[(ii)]  \label{proof_217}
            And indeed we have 
            \begin{align*}
            &   sep-rank\lb \pi \circ f\lb X\rb\rb
            \quad\quad\quad\quad\;\;\;
            \\ &\quad\;\;
                = \underset{i,\, j \in \lc n\rc\times \lc m\rc}{\max}\;\,
               sep-rank\lb \lb \, \pi \circ f\lb X\rb\,\rb_{i,\, j}\rb 
            \\ &\quad\;\;
            = \underset{i,\, j \in \lc n\rc\times \lc m\rc}{\max}\;\, sep-rank\lb 
               \lb\, f\lb X\rb\,\rb_{i,\, j}\rb 
            \\ &\quad\;\;
            = sep-rank\lb f\lb X\rb\rb  
            \end{align*}
            As needed  $\hfill\square$
        \item[(iii)] \label{proof_218}
                The following is decomposition of order $2$
                \begin{align*}
                    &
                    Id\lb X\rb = X = X_A + X_B
                    \quad\quad\quad\quad
                    \quad\quad\quad\quad
                    \quad\quad\quad\quad
                    \\ & 
                    \quad\quad\quad\quad
                    \quad\,
                    =  
                    X_A \odot \mathbf{1}_{n \times m} \,+\, \mathbf{1}_{n \times m} \odot X_B
                \end{align*}
                when $\mathbf{1}_{n\times m}$ is the $n\times m$ matrix with $1$ 
                on all of its entries.
                $\hfill\square$ 
        \item[(iv)] See Subappendix A.1  $\hfill\square$ 
        \end{description}
    \end{adjustwidth}
\end{proof}\label{lem_last_1}

\textbf{MLP-mixer bounds.}  
Relying on the last lemma (\ref{lem_last_1}),
the next lemma presents an upper bound 
on the separation-rank of mixer layer.

\begin{lemma}\label{layer_bound}
    Let $L_{mlp}$, be a general mlp layer as defined in equation (\ref{eq7_2}) 
    \begin{align}
        &
        L_{i,\; m, n}^{2, \,mlp}\lb X\rb 
         = 
        \mathbf{1}_O\lc k\rc \cdot
        \sigma_2\lb W_k^o \pi_o\lb X\rb\rb
        \quad\quad\quad
        \label{eq7_22}
        \\ & \quad\quad\quad
             \quad\quad
        + \mathbf{1}_E\lc k\rc 
        \cdot \sigma_2\lb \pi_e\lb X\rb W_k^e  \rb 
        + \mathbf{1}_R\lc k\rc \pi_r\lb X\rb
        \nonumber
    \end{align}
   
    and let $f: \mathcal{X}^M \to \mathbb{R}^{n \times m}$ be a general function. 
    Then, the following upper bound on the separation rank holds 
    \begin{align}
    &
        sep-rank\lc L_{i,\;k, m, n}^{2, \,mlp}\circ f\lb X\rb\rc 
    \label{eq18_2} 
    \\ & \quad\;\;\;
    \leq\, 
        n^2\,\cdot\, sep-rank\lc f\lb X\rb\rc^2 
        + sep-rank\lc f\lb X\rb\rc    
    \nonumber
    \end{align}
\end{lemma}
\textit{Proof}.$\,$ See Appendix B. $\hfill\square$

Applying the last lemma (\ref{layer_bound}) 
recursively, we may get the following  
upper bound on the separation-rank of 
a full mlp-mixer architecture:

\begin{theorem}
    Let 
    $\;y_{p,\; m,n}^{2, \, mlp}\;: \mathbb{R}^{n \times m} \to \mathbb{R}^{n \times m}$ be an mlp-based architecture  
    with depth $p$ of the form 
    $y_{p,\; m,n}^{2,\, mlp}\lb X\rb \,=\, 
    L_{p,\; m, n}^{2, \,mlp}
    \circ \;.\,.\,.\;\circ 
    L_{1,\; m, n}^{2, \,mlp}
    \lb X\rb
    $, 
    then we have the following bound on the separation-rank
    of the entire model. 
    \begin{align}
        sep-rank\lb \,y_{p,\; m,n}^{2, \,mlp}\,\rb 
        \;\leq\; 
        \lb 2 H \cdot m^2 \,\cdot \, n^2\rb^{2^p}        
    \end{align}
    writing differently, we have 
    \begin{align}
        &
        \log_3\lb\, sep-rank\lc 
        \,y_{p,\; m,n}^{2, \, mlp}
        \,\rc \,\rb 
        \quad\quad\quad\quad\quad\;
        \label{eq__890}\\ & 
        \quad\quad\quad\;
        \quad\quad\quad\;\;
        \;\leq\; \log_3 
        \lb\, 2 H \cdot m^2 
        \,\cdot \, n^2 \;\rb \,
        \cdot\, 2^p        
    \nonumber
    \end{align}
\end{theorem}
\textit{Proof}.$\,$ See Appendix C. $\hfill\square$

\textbf{Transformer bounds.} 
The main thing left for us to do in order to conduct expressiveness comparisons, between the transformer and the mlp-based architectures, is to develop similar lower bounds for attention-based mechanisms, and then show that the found lower bound for the transformer is asymptotically larger than the corresponding upper bound (\ref{eq18_2}) we established thus far.

We will start by presenting an equivalent upper bound for 
transformer architectures for the one we just established 
for the mlp-based architectures (\ref{eq18_2}). 
Getting such an upper bound is useful in order to show 
the tightness of our lower bound. Such tightness results  
would mean that our expressiveness gap result
could not be widened by achieving a better lower 
bound for the transformer and that if someone manages to show 
that the upper bound for the mlp-mixer is tight, 
then at least under the relaxed model's assumptions  
the gap we found is exact in the sense that no larger gap 
exists.

\begin{theorem}\label{229}
    Let 
    $y_{p, H}^{R}:
    M_{n \times m}\lb \mathbb{R}\rb 
    \to M_{n \times m}\lb \mathbb{R}\rb$, 
    be transformer architecture   
    without activations and normalization layers   
    with depth $p$ and residual connections  
    of the form 
    \equ{
        y_{p, H}^{R}\lb X\rb = L_{p, H}^{R} \circ ... \circ L_{1, H}^{R}\lb X\rb 
    }
    with all the layers and matrices having the same dimensions,
    $W_p^{i j} \in M_{m \times n}\lb \mathbb{R}\rb\,$.     
    Then, the following bound on the separation rank holds
    \equ{\label{eq730}
        sep-rank\lb y_{p, H}^{R}\rb
        \;\leq\;
        \lb 2 H \cdot m^2 \,\cdot \, n^2\rb^{3^p}
    }    
\end{theorem}
\textit{Proof}.$\,$ See Appendix D. $\hfill\square$

Finally, to get a lower bound, we are relying on theorem (7.1) from the book (\cite{levine2022tensors}) to get that for linear transformers
without residual connections the following theorem 
holds 

\begin{figure*}[t]
\begin{center}
    \includegraphics[width=.99\columnwidth]
    {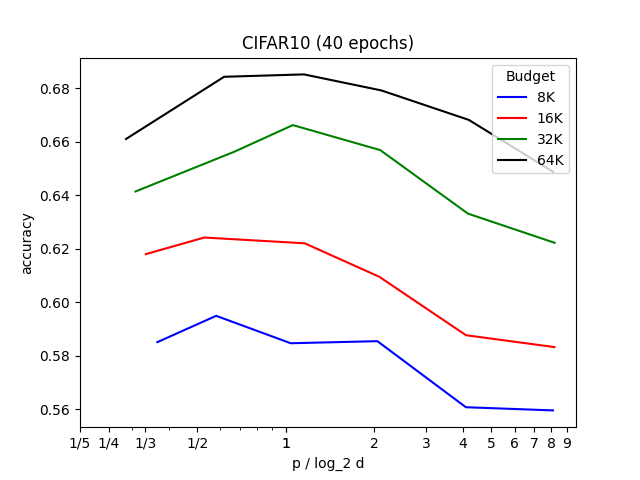}
    \hfill
    \includegraphics[width=.99\columnwidth]
    {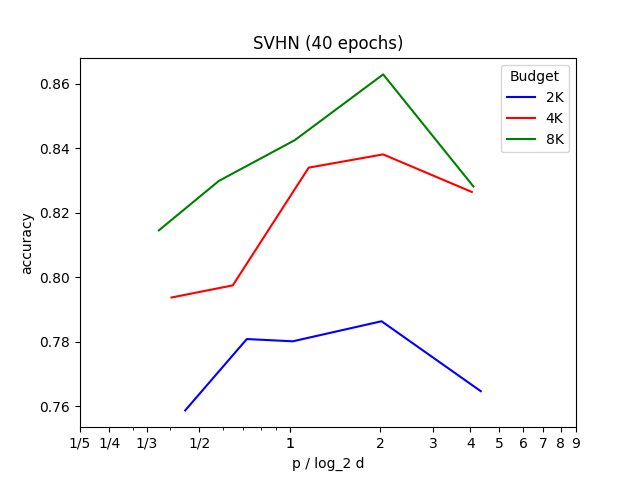}
    \hfill
    \includegraphics[width=.99\columnwidth]
    {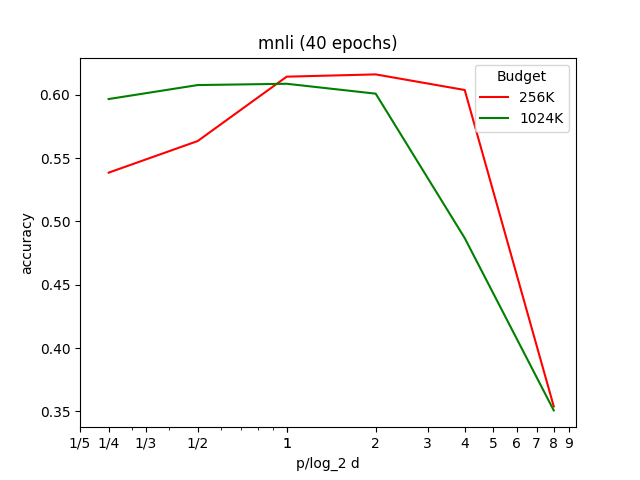}
\end{center}
\caption{Mixer Performance for different depth-to-width ratios. Results obtained on the CIFAR10, SVHN and MNLI datasets when trained for $40$ epochs, with $9$ different budgets averaged over $5$ seeds. We use smaller budgets for easier datasets. Only runs with a standard deviation smaller than $0.15$ are reported. We can see that the best performance is obtained 
for $1<\frac{p}{\log_2 d}<2$.}
\label{fig:abe_adv_eps}
\vskip -0.2in
\end{figure*}

\begin{theorem}\label{thm_66_2}
    For $p < \log_3 m\,$ there is a weights assignment such that our upper bound 
    \begin{align}
        &
        \log_3\; sep-rank\lb y_{p, H}^{R}\rb
        \label{eq860}\\ & \quad\quad\quad\quad\quad
        \;\leq\;
        3^p \cdot 
        \lc \log_3 \lb 2 H\rb  + 2\log_3 m \,+ \, 2 \log_3 n \rc
        \nonumber
    \end{align}
    is asymptotically tight in the sense 
    \begin{align}
        &
        \log_3 sep-rank\lb y_{p, H}^{R}\rb
        \quad\quad\quad\quad\quad\;
        \label{eq__860}\\ & 
        \quad\quad\quad
        \quad\quad\;\;\;
        \;\geq\;
        3^{p-2}\lc
        \log_3 \lb m-H\rb -p+ 2-\log_3  2\rc        
        \nonumber
    \end{align}
\end{theorem}
\textit{Proof}.$\,$ See Appendix E. $\hfill\square$

\textbf{Results.} 
Comparing the obtained bounds,   
we may end with the following 
theorems regarding the 
expressive gap between the 
transformer and the mlp-mixer architectures.

\begin{conclusion}\label{c1_2}
    We got that for 
    $\mathcal{F}_{B}^{T},\, \mathcal{F}_{B}^{mlp}$ 
    the classes of transformer and mlp-based architectures  
    with up to $B$ parameters 
    respectively. 
    It holds that    
    $\frac{\log_3 sep-rank\lb \mathcal{F}_{B}^{T}\rb}
    {\log_3 sep-rank\lb \mathcal{F}_{B}^{mlp}\rb} \,=\,
    \Omega\lb\, \lb \frac{3}{2}\rb^p \,\rb$. 
    More formally, there is a 
    monotonicity relation of the form
    $\mathcal{F}_{B}^{mlp} \prec_{\lb \frac{3}{2}\rb^p} \mathcal{F}_{B}^{T}$.
\end{conclusion}
\textit{Proof}.$\,$ See Appendix F. $\hfill\square$

\begin{conclusion}\label{c2_2}
    For $p < \log_3 m$ and assuming 
    $p >> \log_3\log_3 m\;$ $n < m^2$,
    $H < \frac{m}{2}$ and $p \geq 13\,$. 
    Then, every mlp-based architecture
    has a strictly smaller expressive power in modeling multi-variable 
    dependencies than any attention-based architecture, 
    when fixing the depth and the parameters budget.
    Also, for $\log_3 m < p < \log_2 m$,  
    then still, transformers enjoy strictly higher expressive power than mlp-based architectures for large enough $p$, 
    and when moving into the depth efficiency regime 
    $p < \log_3 m$ 
    the gap becomes asymptotically exponential in $p$.    
\end{conclusion}

\textit{Proof}.$\,$ See Appendix G. $\hfill\square$

\begin{remark}
    The difference between the last two conclusions is that the first conclusion (\ref{c1_2}) states that the wisest choice of transformer architecture is better than the wisest choice of mlp-architecture, whereas the second conclusion (\ref{c2_2}) states that every transformer with a good depth-to-width ratio is superior to every mlp-based architecture.
\end{remark}

\begin{proposition}\label{p1_2}
    Conclusion (\ref{c1_2}) states dominance relation 
    between transformer and mlp classes with the same depth. 
    When comparing classes of different depth
    $\mathcal{F}_{B,\, p_{n, mlp}}^{mlp},
    \;\mathcal{F}_{B,\, p_{n, T}}^{T}$
    then as long as 
    $\alpha = \underset{n\to \infty}{\limsup}
    \;\frac{p_{n, mlp}}{p_{n, T}} 
    < \log_2 3 \simeq 1.584\,$     
     the following dominance relation still holds  
    $\mathcal{F}_{B, p_m}^{mlp} \prec_{\lb \frac{3}{2^\alpha}\rb^p} \mathcal{F}_{B, p_t}^{T}$.    
\end{proposition} 

\textit{Proof}.$\,$ See Appendix H. $\hfill\square$


\begin{remark}
   The last (\ref{p1_2}) proposition 
   leaves open the possibility 
   that if someone can scale mlp-architectures
   $\simeq 1.58$ deeper   
   than transformer architectures 
   while using the same budget, 
   then it may be possible that the mlp-architectures
   would have a higher ability to model multi-variable dependencies.
   However, our upper bound over the separation rank of mlp architectures
   is not necessarily tight, so we did not claim it but we leave this possibility open for further research. 
\end{remark}


\section{Experiments}
To assess our theory we derived a few predictions from it and asses them in experiments as shown below. 
The first experiment 
is also intended to support the 
$\sigma_2$ relaxation performed above (\ref{mlp_relax}), 
by using the separation-rank of the relaxed 
MLP-mixer to predict the optimal depth-to-width 
ratio for the MLP-mixer model 
and assessing it by experiments. 
\subsection{Depth to width ratio}\label{exp_1}
Our first prediction is about the optimal depth-to-width ratio for the mixer architecture, 
when coming to this issue, then for the transformer architectures
as shown in the appendices and relying on
 \cite{levine2022tensors, levine2020limits}
 it holds that the optimal depth to width ratio for transformers 
 architectures is $p\approx\log_3 d$ 
 where $p$ and $d$ denote the transformer depth and width respectively.

In general, as shown in appendixes (I) for every architecture 
$y_{p, d}$ with 
\equ{
    \log_\alpha\lc\, sep-rank\lb y_{p, d}\rb\,\rc 
    = \Theta\lb \, Q_1\lb p, d\rb \cdot \alpha^{p}\,\rb
}
for $p < \log_\alpha d$ and 
\equ{
    \log_\alpha\lc\, sep-rank\lb y_{p, d}\rb\,\rc 
    = \Theta\lb \, Q_2\lb p, d\rb \,\rb
}
for $p> \log_\alpha d$  
where $Q_1, \, Q_2:\mathbb{N}^2 \to \mathbb{N}$ is some multinomial with a finite degree, 
and $1 < \alpha \in \mathbb{R}$ is the exponent basis when fixing a budget $B$ 
the optimal depth to width ratio satisfies
$1<\frac{p}{\log_\alpha d}$ and hence in the mixer case since we manage to show that 
\equ{
    \log_2\lc\, sep-rank\lb y_{p, d}\rb\,\rc 
    = O\lb \, Q_1\lb p, d\rb \cdot 2^{p}\,\rb
}
for $p < \log_\alpha d$ and 
\equ{
    \log_2 \lc\, sep-rank\lb y_{p, d}\rb\,\rc 
    = O\lb \, Q_2\lb p, d\rb \,\rb
}
but we did not show the appropriate lower bound then we may hypothesize that for the mixer it also holds 
$p^* = \log_{\alpha_{mixer}} d^*$
 when
 $1 < \alpha_{mixer} < 2$ and in particular 
 \equ{
    2 = \alpha_{mixer} < \alpha_{transformer} = 3
 }
 \graphicspath{{./Experiment-Results}}
 We tested this hypothesis by examining the accuracy of multiple different models with the same parameter budget, but with different depth-to-width ratios on the CIFAR10, SVHN and MNLI datasets when trained for 40 epochs.

 As we can see (\ref{fig:abe_adv_eps}) 
 the pick performance 
 is obtained for 
 \equ{
    1 < \frac{p}{\log_2 d} < 2
 }
 and note that 
 \equ{
    \frac{p}{\log_3 d}
    = \frac{p}{\log_2 d} \cdot \log_2 3 \approx 1.58 \cdot \frac{p}{\log_2 d}
 }
 hence for the transformer it just holds that 
 \equ{
    \frac{p_\textit{transformer}}{\log_2 d_\textit{transformer}}
    \approx \frac{5}{8} < 1 < \frac{p_{mixer}}{\log_2 d_{mixer}}
 }

\subsection{Data-size and training-time}

It has been shown by Li et al. \cite{li2020train}   
deeper RoBERTa models tend to converge faster, 
\cite{li2020train} also shows that 
larger and more expressive models  
usually converge faster, unless there are overfitting issues. 
Their results asses our theory about the larger effective depth 
of the transformer architectures    
relative to the mlp-based ones,     
which results in slower convergence of the mixer models as indicated 
by \cite{tolstikhin2021mlp}.

\section{Conclusions and discussion}
To conclude, 
we showed the existence of an exponential gap 
in the expressive power  
between MLP-based architectures and attention-based ones   
in their ability to model multi-variable dependencies. 
This may explain the performance gap in vision tasks
as well as the nonexistence of mlp-based architectures 
for NLP tasks. 
This also suggests  
that mlp-based architectures are indeed inferior   
to the attention-based ones and that although permutations-based strategies
may give some improvements    
they may not suffice to close the gap  
since those architectures have degraded expressive power   
in the sense of modeling long dependencies.
We also showed that this gap sustains as long as the mlp-architecture, 
with the same budget, is not $1.58$ times deeper.
However, we leave the question open, of how much depth increase is required for the mlp to achieve the same expressive power as the transformer. 
Say it differently, the transformer can achieve
a larger effective depth using fewer layers relative to the mlp.
This suggests some more explanation for the wide success 
of the attention-based mechanisms for various different tasks.

\newpage
\textbf{Acknowledgements} 
The first author would like to thank Noam Weiss for 
useful conversations. 
This work was partially supported by the 
European Research Council (ERC) under the European Unions Horizon 2020 research and innovation programme (grant ERC CoG No.863839).

{\small
\bibliographystyle{ieee_fullname}
\bibliography{main.bib}
}

\end{document}